\documentclass[runningheads]{llncs}
\usepackage[T1]{fontenc}
\usepackage{graphicx}

\usepackage{algorithm}
\usepackage{algorithmicx}
\usepackage{algpseudocode}
\usepackage[thicklines]{cancel}

\usepackage{bm}
\usepackage{ulem}

\usepackage{multirow}
\usepackage{booktabs}
\usepackage{array}
\usepackage{subfigure} 
\usepackage{amsmath}
\usepackage{amssymb}
\usepackage{ulem} 
\usepackage{threeparttable}
\usepackage{makecell}
\usepackage{longtable}
\usepackage{hyperref}
\usepackage[misc]{ifsym}
\usepackage{bbding}

\usepackage{xcolor}

\begin{document}
\title{Evolution-based Feature Selection for Predicting Dissolved Oxygen Concentrations in Lakes}
\titlerunning{Evolution-based Feature Selection for Predicting Lake DO Concentrations}

\author{Runlong Yu\inst{1}, Robert Ladwig\inst{2}, Xiang Xu\inst{3}, Peijun Zhu\inst{4}, Paul C. Hanson\inst{5}, Yiqun Xie\inst{6}, \and Xiaowei Jia\inst{1}\textsuperscript{(\Letter)}}

\authorrunning{R. Yu et al.}

\institute{University of Pittsburgh, Pittsburgh, USA 
	\and Aarhus University, Aarhus, Denmark 
	\and University of Science and Technology of China, Hefei, China 
	\and Georgia Institute of Technology, Atlanta, USA 
	\and University of Wisconsin-Madison, Madison, USA 
	\and University of Maryland, College Park, USA \\
	\email{{ruy59,xiaowei}@pitt.edu, rladwig@ecos.au.dk, demon@mail.ustc.edu.cn, zpj@gatech.edu, pchanson@wisc.edu, xie@umd.edu}
}

\maketitle              

\begin{abstract}
	
	Accurate prediction of dissolved oxygen (DO) concentrations in lakes requires a comprehensive study of phenological patterns across ecosystems, highlighting the need for precise selection of interactions amongst external factors and internal physical–chemical–biological variables. This paper presents the \textsl{Multi-population Cognitive Evolutionary Search (MCES)}, a novel evolutionary algorithm for complex feature interaction selection problems. MCES allows models within every population to evolve adaptively, selecting relevant feature interactions for different lake types and tasks. Evaluated on diverse lakes in the Midwestern USA, MCES not only consistently produces accurate predictions with few observed labels but also, through gene maps of models, reveals sophisticated phenological patterns of different lake types, embodying the innovative concept of  ``AI from nature, for nature''.

\end{abstract}

\section{Introduction}

The concentration of dissolved oxygen (DO) in lakes is a key indicator of water quality and the health of freshwater ecosystems.  
Effective DO monitoring is critical for sustaining aquatic biodiversity and ensuring water security for human consumption~\cite{wilson2010water}. DO concentrations are influenced not only by the exchange of oxygen between air and water, but also by the metabolic processes of primary production and respiration~\cite{sommer2012beyond}. As articulated by Edward A. Birge one century ago~\cite{birge1906gases}: The fluctuations in a lake's oxygen illustrate its ``life cycle'' more clearly than many other ecological indicators. This is particularly evident in nutrient-rich eutrophic lakes, where algal blooms can significantly deplete oxygen, creating detrimental ``dead zones'' for aquatic life.

Accurate prediction of DO concentrations requires a comprehensive study of different phenological patterns across various ecosystems. 
In particular, DO concentration is closely intertwined with ecosystem phenology, influenced by morphometric and geographic information, mass fluxes, weather conditions, trophic state, and watershed land use. In deeper lakes, for instance, light scarcity and decreased mixing with the oxygen-rich surface can lower oxygen~\cite{solomon2013ecosystem,phillips2020time}. Temperature fluctuations impact oxygen solubility and biochemical activities~\cite{staehr2010lake}. Land use changes reshape DO patterns and metabolism phenology~\cite{jenny2016urban,woolway2021phenological}.

Given the importance of predicting DO concentration, scientists across limnology, hydrology, meteorology, and environmental engineering have devised process-based models to simulate the dynamics of freshwater ecosystems. These models, aimed at evaluating the effects of external and internal factors, often combine hydrodynamic and water quality models~\cite{janssen2015exploring}. Examples include DYRESM~\cite{hamilton1997prediction}, GLM~\cite{hipsey2019general}, MyLake~\cite{saloranta2007mylake}.
These models utilize first-order principles (e.g., mass and energy conservation), but also involve many parameterizations or approximations due to incomplete physical knowledge or excessive complexity in modeling certain complex processes, resulting in inherent model~bias.

Advanced data-driven methods like deep learning~\cite{lecun2015deep},  offer an alternative to process-based models for complex scientific problems (e.g., prediction of DO concentration). 
Their success is contingent upon effective feature selection~\cite{correia2013feature,li2017feature,brookhouse2022fair}, however, most extant methods face several major challenges in this problem.  
Firstly, predicting DO involves sophisticated phenological patterns across various metabolic processes. Directly considering a large number of feature interactions can easily introduce noise, while manual selection risks overlooking critical details. 
Secondly, most feature selection approaches rely on global models built under expert guidance and lack the flexibility to adapt to various tasks and datasets. As a result, they often fail to select relevant feature interactions for different lake types and prediction tasks. 
Finally, the sparse nature of DO data further complicates model training, as frequent and comprehensive data collection is hindered by substantial human labor and material costs.

To address these challenges, cognitive evolutionary search (CELS) has been developed~\cite{yu2023cognitive}, utilizing an evolutionary algorithm to adaptively evolve models for selecting feature interactions under task-specific guidance. Unlike conventional methods constrained by model fitness evaluation~\cite{zhang2019two,zhao2021coea,xue2015survey,telikani2021evolutionary}, CELS introduces a novel model fitness diagnosis technique. Despite its advancements, CELS still contends with issues of ``reproductive isolation'', which may result in the convergence of similar models within a population, thereby limiting the development of specialized models for distinct tasks~\cite{tang2016negatively,li2016seeking}.

Leveraging insights from CELS, this paper proposes an advanced evolutionary algorithm, namely \textsl{Multi-population Cognitive Evolutionary Search (MCES)}, where cognitive ability refers to the malleability of organisms to orientate to the environment~\cite{yu2023cognitive}. MCES utilizes multi-population models to effectively serve diverse lake types and predictive tasks, mirroring the adaptive strategies of species in diverse habitats. In MCES, feature interactions are envisaged as genomes and models as organisms within ecosystems, conceptualizing tasks as the natural environments these organisms inhabit.
Internally, MCES assesses the capacities of genes, with mutations occurring probabilistically when existing traits are detrimental to survival. The models within MCES undergo crossover and mutation within their respective populations and, though less commonly, engage in inter-population crossover. This genetic diversity and flexibility allow models to dynamically adapt and select appropriate phenological feature interactions, catering to specific environmental conditions of different lake types and tasks.

\section{Related Work}

Existing research suggests that evolution-based feature selection is often limited to filters and wrappers due to model fitness evaluation constraints~\cite{xue2015survey,telikani2021evolutionary}. Specifically, wrapper methods use the performance of the learning algorithm as its evaluation criterion, while filter methods use the intrinsic characteristics of the data. On the other hand, embedded approaches simultaneously select features and learn a classifier, therefore conventional algorithms cannot evaluate the fitness of the model~\cite{xue2015survey,telikani2021evolutionary}. Only genetic programming (GP) and learning classifier systems (LCSs) are able to perform embedded feature selection, but they are not practical~\cite{guyon2003introduction,muni2006genetic,lin2008classifier,xue2015survey,telikani2021evolutionary}. For additional research on evolution-based feature selection, please refer to~\cite{yu2023cognitive}.

Research on the population structure of EAs has demonstrated the benefits of segmenting the initial population into multiple sub-populations. These sub-populations exchange information, and regrouping operators are triggered at regular intervals to maintain the population's diversity and balance exploitation with exploration capabilities~\cite{wu2016differential,zhu2020self}.
Various models like the shuffle or update parallel differential evolution (SOUPDE)~\cite{weber2011shuffle} and the multi-population-based cooperative coevolutionary algorithm (MPCCA)~\cite{shang2014multi} utilize unique mutation strategies and population dynamics to optimize performance across diverse problem-solving scenarios. 

\section{Problem Definition and Preliminaries}

\subsubsection{Problem definition.} Our goal is to predict the DO concentration at a daily scale. We simplify our analysis by dividing the water column into two distinct layers with separate oxygen and metabolic kinetics: the epilimnion (upper surface layer) and the hypolimnion (lower bottom layer). We treat the DO prediction for the epilimnion and hypolimnion as two tasks. 

For each lake, we have access to its phenological features $\pmb{x}_t$  on each date $t$. 
These features, spanning $m$ diverse fields $\pmb{x}_t = \{ x_t^1, \cdots, x_t^m \}$, encompass morphometric and geographic details such as lake area, depth, and shape; flux-related data like ecosystem and sedimentation fluxes; weather factors comprising wind speed and temperature; a range of trophic states from dystrophic to eutrophic; and diverse land use proportions extending from forests to wetlands. 
Besides input features, we also have observed DO concentrations $y_t$ (for both the epilimnion $y_{t}^{\rm epi}$ and hypolimnion $y_{t}^{\rm hyp}$) on certain days.

We use an embedding layer to convert input phenological features into a series of multi-field feature embeddings $\pmb{f}_t = [\pmb{f}^1_t, \cdots, \pmb{f}^m_t]$, where $\pmb{f}^i_t = \text{embed}(x^i_t)$. Initially, numerical features are bucketed into categories, and each category is then represented as a one-hot vector that is transformed into an embedding vector through a perceptron layer~\cite{yu2021xcrossnet}. Our model uses these embeddings to predict DO concentrations $\hat{y}_t$ for both the epilimnion $\hat{y}_{t}^{\rm epi}$ and hypolimnion $\hat{y}_{t}^{\rm hyp}$.

\subsubsection{Feature interaction selection.}

The process of feature interaction selection aims to identify the most informative feature interactions that can facilitate the prediction of target DO concentrations $\mathcal{H}: \mathcal{M}(\pmb{f}, \pmb{g}(\pmb{f}))\rightarrow \hat{y}$,  where $\pmb{g}$ denotes the set of operations to interact on feature pairs, and $\pmb{g}(\pmb{f})$ denotes the set of interactions.
In DO prediction case, the algorithm $\mathcal{H}$ aims to minimize the MSE loss for the outputs of the prediction model $\mathcal{M}$, given as: 
\begin{equation} 
	\label{eq0}
	\mathcal{L}(\mathcal{M}) = \frac{1}{|B|}\sum_{t \in B} \Big(y_{t} -\hat{y}_{t} \Big) ^2, 
\end{equation} where $B$ denotes the set of instance indices in a mini-batch, $\hat{y}_{t}$ denotes the predictive result given through the learned model.

As the fundamental components in feature interaction, operations are functions where two individual features are converted into an interaction. For the sake of simplicity, we adopt four representative operations as candidate operations to present instantiations of MCES, i.e., $\pmb{g}=\{ \oplus, \otimes,  \boxplus, \boxtimes \}$. Detailed explanations of these operations and their use in evolution-based feature selection are provided in \href{https://zenodo.org/doi/10.5281/zenodo.10993058}{\textbf{Appendix A}}.

\subsubsection{Metabolic process-based model. }

In this work, we use a metabolic process-based model to simulate DO labels~\cite{ladwig2022long}. The process-based model divides the water column into the upper epilimnion and lower hypolimnion during stratified periods. It is focused on analyzing metabolic dynamics in warmer months. Flux features (denoted as $F$) affecting DO concentrations are calibrated with observed data~\cite{ladwig2022long}. 
More details of the process-based model are provided in \href{https://zenodo.org/doi/10.5281/zenodo.10993058}{\textbf{Appendix~B}}.

\section{Overall Framework}
\begin{figure} [!t]
	\includegraphics[width=\linewidth]{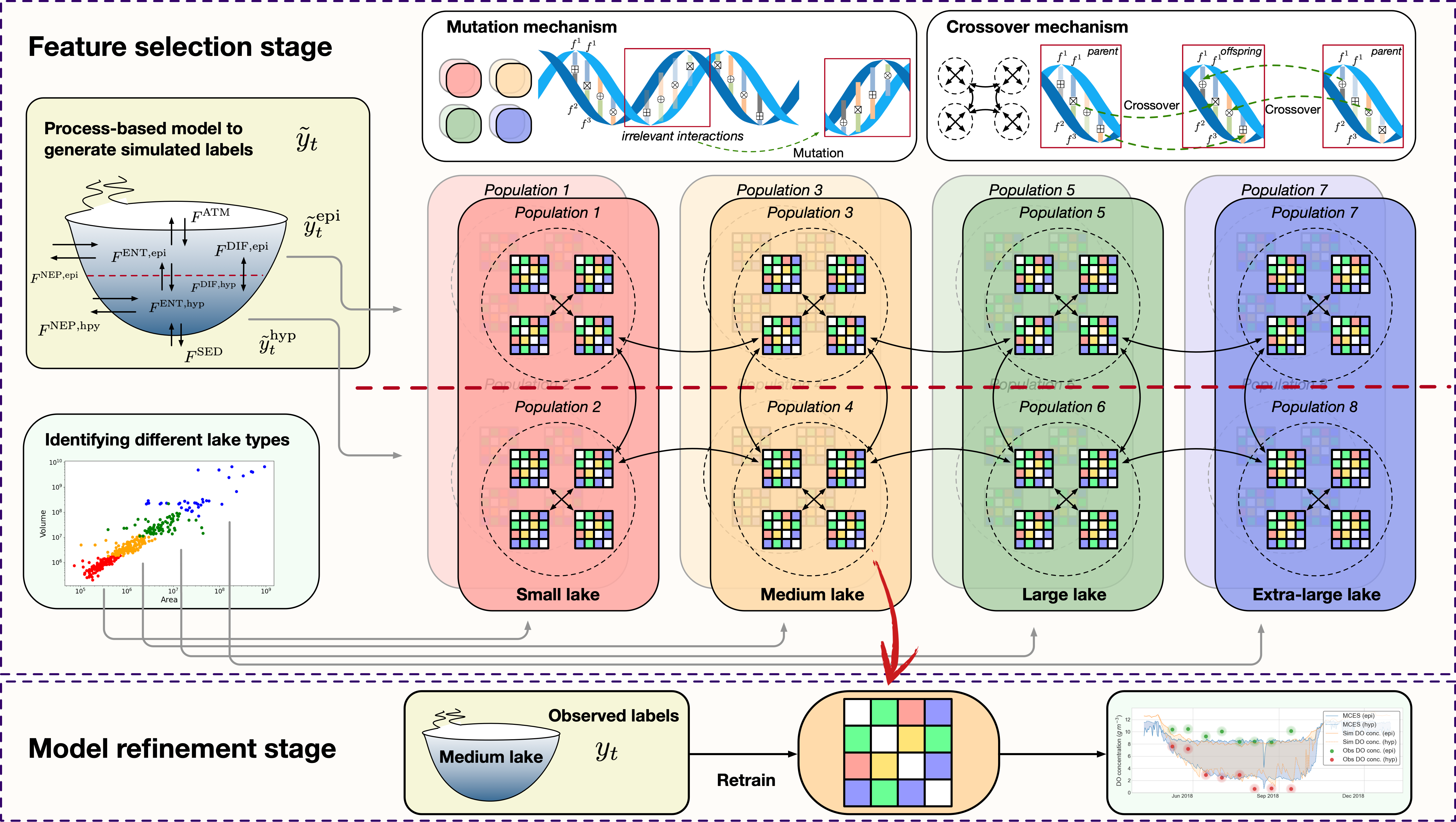}
	\caption{ Overall framework.  } 
	\label{fig:2}
\end{figure}

The overall framework is depicted in Figure~\ref{fig:2} and involves two stages of learning: (1) MCES for feature interaction selection using simulated labels, and (2) model refinement using real observed labels.

Specifically, we first utilize metabolic process-based models to generate simulated DO labels given the phenological features $\pmb{x}_t$. Using these simulated labels, we implement our proposed feature interaction selection algorithm MCES. The models in MCES adaptively evolve to select relevant feature interactions within populations for different lake types and tasks.
In the subsequent stage, we refine these evolved models using real-world observed DO concentration data. 
This mirrors natural genetic decoding, where selected feature interactions are further optimized to reflect actual ecological dynamics.

\subsubsection{Identifying different lake types.} 
As a preprocessing step, lakes are classified based on characteristics critical to oxygen dynamics—primarily surface area and volume. Larger surface areas improve atmospheric oxygen exchange, while greater volumes are associated with higher oxygen consumption in deeper waters. We use a balanced K-means clustering algorithm to uniformly distribute the lakes in dataset $\pmb{L}$ into four categories based on volume and area: small lakes $\pmb{L}_{\rm S}$, medium lakes $\pmb{L}_{\rm M}$, large lakes $\pmb{L}_{\rm L}$, and extra-large lakes $\pmb{L}_{\rm xL}$.


\section{Multi-population Cognitive Evolutionary Search}

In this section, we detail the feature selection stage, employing simulated labels $\tilde{y}_{t}$ to implement MCES. We begin by introducing the establishment of multiple populations, and explain how each individual within these populations is modeled. We then delve into the fundamental mutation and crossover mechanisms. Lastly, we provide an instantiation of the search process.

\subsection{Multiple Populations}

In MCES, each population comprises a set of models designed specifically for a particular combination of lake type and prediction task (e.g., predicting DO in the epilimnion of large lakes). Consequently, we  establish eight unique populations: $\pmb{P}^{\rm epi}_{\rm S}$, $\pmb{P}^{\rm hyp}_{\rm S}$, $\pmb{P}^{\rm epi}_{\rm M}$, $\pmb{P}^{\rm hyp}_{\rm M}$, $\pmb{P}^{\rm epi}_{\rm L}$, $\pmb{P}^{\rm hyp}_{\rm L}$, $\pmb{P}^{\rm epi}_{\rm xL}$, $\pmb{P}^{\rm hyp}_{\rm xL}$. Here the subscript $\{\rm S,M,L,xL\}$ represents different lake types, and the superscript $\{\rm epi, hyp\}$ represents two different prediction tasks (i.e., epilimnion or hypolimnion).

With a population size of $n$ (where $n>1$), we initialize each population with $n$ models for feature interaction selection: $\pmb{P} = \{ \mathcal{M}_1, \cdots, \mathcal{M}_n \}$. We conceptualize each model as a natural organism that evolves to enhance its traits for improved fitness within its environment. These traits, inherited through the organism's genomes, stem from the interplay between features and operations, much like nucleotides and their connections. Following various linkages of nucleotides, we extend the operation set with four types of operations as the search space, i.e.,  $\pmb{g}=\{ \oplus, \otimes,  \boxplus, \boxtimes \}$. If $g_k$ is a chosen operation from $\pmb{g}$, an interaction $g_k(\pmb{f}^i_t, \pmb{f}^j_t)$ is defined by applying the operation $g_k$ to a pair of features $(\pmb{f}^i_t, \pmb{f}^j_t)$.

Motivated by the goal of enhancing model fitness through the preservation of beneficial genetic information, we aim to discern and prioritize important features and their interactions via a parameterized method, called internal genetic evaluation. The idea is to introduce a set of relevance parameters to strengthen relevant feature interactions while diminishing or mutating those that contribute less.  
In this context, we define relevance parameters  for features $\pmb{f}_t $ and interactions $\tilde{g}(\pmb{f}_t)$ as $\pmb{\alpha} = \{ \alpha_i | 1\leqslant i\leqslant m\} $ and $\pmb{\beta} =\{ \beta_{i,j} | 1\leqslant i<j\leqslant m \} $, respectively. Here, $\tilde{g}(\pmb{f}_t)$ denotes the interaction of applying any operations from $\pmb{g}$ to a pair of features.
The predictive response of our model at time step $t$ is formulated as: 
\begin{equation}   \label{eq1}
	\hat{y}_{t} = \mathcal{M}\big( \pmb{\alpha} \cdot \pmb{f}_t,  \pmb{\beta} \cdot \tilde{g}(\pmb{f}_t) \big),
\end{equation} where $\mathcal{M}$ can be any individual model in the population. In this work, we use a sequence encoder with Long-Short Term Memory (LSTM) networks~\cite{hochreiter1997long} to efficiently encode temporal information and the dynamics of feature interactions. The model $\mathcal{M}$ is thus depicted as:
\begin{equation}   \label{eq2}
	\begin{split}
		& \pmb{h}^{\iota}_t = {\rm LSTM} \Big([ \pmb{\alpha} \cdot \pmb{f}_t,  \pmb{\beta} \cdot \tilde{g}(\pmb{f}_t)];\pmb{h}^{\iota}_{t-1} \Big), \\
		& \hat{y}_{t} = \pmb{W}^{\iota} \cdot \pmb{h}^{\iota}_t + \pmb{b}^{\iota},
	\end{split} 
\end{equation} where $\pmb{h}^{\iota}_t$ represents a series of hidden states, and $\pmb{W}^{\iota}$ and $\pmb{b}^{\iota}$ denote the weight and bias parameters, respectively. The loss function for model $\mathcal{M}$, calculated using simulated labels $\tilde{y}_{t}$, is defined~as: 
\begin{equation}  
	\label{eq3}
	\mathcal{L}(\mathcal{M}) = \frac{1}{|B|}\sum_{t \in B} \Big(\tilde{y}_{t} -\hat{y}_{t} \Big) ^2, 
\end{equation} where $B$ denotes the set of instance indices within a mini-batch.

We use a regularized dual averaging (RDA) optimizer to learn the relevance parameters $\pmb{\alpha}$ and $\pmb{\beta}$, with the aim to distinguish between relevant and irrelevant feature interactions through this process~\cite{xiao2009dual,chao2019generalization}. When the absolute value of the cumulative gradient average value in a certain position in $\pmb{\alpha}$  or  $\pmb{\beta}$  is less than a threshold, the weight of that position in relevance parameters will be set to 0, resulting in the sparsity of the relevance~\cite{xiao2009dual,liu2020autofis}. Meanwhile, the parameters of the embedding layer (for $\pmb{f}_t$) are optimized using the Adam optimizer~\cite{kingma2014adam}. 
Unlike AutoML~\cite{liudarts,liu2021survey}, which categorizes $\pmb{\alpha}$ and $\pmb{\beta}$ as high-level decisions and treats feature embeddings as lower-level variables for bi-level optimization, our approach simplifies this process. 
To circumvent the complex and costly bi-level optimization, we update feature embeddings and relevance parameters jointly using single-level optimization with gradients  on the training set, as
$\nabla_{\pmb{f}} \mathcal{L}(\pmb{f}_{iter-1}, \pmb{\alpha}_{iter-1}, \pmb{\beta}_{iter-1}) $ and $\nabla_{\pmb{\alpha,\beta}} \mathcal{L}(\pmb{f}_{iter-1}, \pmb{\alpha}_{iter-1}, \pmb{\beta}_{iter-1})$, respectively.

\subsection{Mutation Mechanism and Crossover Mechanism}

With our definitions of population and feature interaction selection models, we further detail the mutation and crossover mechanisms in MCES. The crossover mechanism is bifurcated into intra-population and inter-population crossover.

\subsubsection{Mutation mechanism.} The mutation serves as a fundamental mechanism of our search process. It primarily aims at mutating the operations associated with irrelevant interactions into alternative operations, and thus generating a new model (the offspring). 
Specifically, for an interaction $g_k(f^i_t, f^j_t)$, mutation is triggered with a probability $\sigma$ after every $\tau$ steps if the relevance parameter $\beta_{i,j}$ drops below a threshold $\lambda$. In other words, 
to regenerate a new interaction, the operation $g_k$ of the interaction $g_k(f^i_t, f^j_t)$ mutates into another operation $g_l $, given as: 
\begin{equation} 
	\label{eq5}
	g_k = \left\{
	\begin{aligned}
		g_l  \;  \text{with probability} \; \sigma, \;\;\;\;  & \text{if} \quad  \beta_{i,j} < \lambda, \\
		g_k, \;\;\;\;\;\;\;\;\;\;\;\;\;\;\;\;\;\; & \text{otherwise},
	\end{aligned}
	\right.
\end{equation}
where $g_l$ is randomly selected from the operation set as $g_l = \{ g\, | \, g \in \pmb{g}, g \neq g_k \}$. The new interaction $g_l(f^i_t, f^j_t)$ replaces the irrelevant interaction $g_k(f^i_t, f^j_t)$, and its corresponding relevance $\beta_{i,j}$ is reset. Consequently, the parent model $\mathcal{M}$ evolves into its offspring $\mathcal{M}'$, which incorporates  these fresh interactions with revised relevance $\pmb{\beta}'$, and maintains features with relevance $\pmb{\alpha}'$ inherited from $\pmb{\alpha}$.

\subsubsection{Intra-population crossover mechanism.} 
Given a population $\pmb{P} = \{ \mathcal{M}_1, \cdots, $ $\mathcal{M}_{\nu}, \cdots, \mathcal{M}_n \}$, 
we use $\pmb{\beta}^{\mathcal{M}_{\nu}}$ to denote the relevance parameters of interactions for each model $\mathcal{M}_{\nu}$. The obtained $\pmb{\beta}^{\mathcal{M}_{\nu}}$ can vary across different models in $\pmb{P}$.  Therefore, within this population, the models may have a variety of operations for interacting with each feature pair $(f^i_t, f^j_t)$, represented as  $g_{i,j}^{\pmb{P}} = \{ g_{i,j}^{\mathcal{M}_1}, \cdots,  g_{i,j}^{\mathcal{M}_\nu}, \cdots,  g_{i,j}^{\mathcal{M}_n} \}$. 
The intra-population crossover mechanism aims to select the most suitable operation (of which interaction has the largest relevance) within the population to apply on the feature pair for the offspring model $\mathcal{M}'$, given as:   
\begin{equation}  
	\label{eq6}
	g_{i,j}^{\mathcal{M}'} = \arg \max_{g_{i,j}^{\mathcal{M}_\nu} \in \, g_{i,j}^{\pmb{P}}} \beta_{i,j}^{\mathcal{M}_{\nu}}.
\end{equation} Meanwhile, the relevance parameters of interactions in this offspring model are inherited from their respective parent models (i.e., the selected $\mathcal{M}_\nu$).

\subsubsection{Inter-population crossover mechanism.} For two distinct populations $\pmb{P}_A = \{ \mathcal{M}^A_1, \cdots, \mathcal{M}^A_{\nu}, \cdots, \mathcal{M}^A_n \}$ and $\pmb{P}_B = \{ \mathcal{M}^B_1, \cdots, \mathcal{M}^B_{\nu}, $ $\cdots, \mathcal{M}^B_n \}$, we use $\pmb{\beta}^{\mathcal{M}^A_{\nu}}$ and $\pmb{\beta}^{\mathcal{M}^B_{\nu}}$ to denote the relevance of interactions for two populations $\pmb{P}_A$ and $\pmb{P}_B$, respectively. The inter-population crossover mechanism works as follows: For each feature pair $(f^i_t, f^j_t)$, we select the most suitable operation from $\pmb{P}_B$ to interact on the feature pair in the offspring model of $\pmb{P}_A$. Conversely,
the most suitable operation of the feature pair from $\pmb{P}_A$ is selected for the offspring model of $\pmb{P}_B$, given as:
\begin{equation} \label{eq7}
	g_{i,j}^{\mathcal{M}_A'} = \arg \max_{g_{i,j}^{\mathcal{M}^B_\nu} \in \, g_{i,j}^{\pmb{P}_B}} \beta_{i,j}^{\mathcal{M}^B_{\nu}}, \;\;\;\;
	g_{i,j}^{\mathcal{M}_B'} = \arg \max_{g_{i,j}^{\mathcal{M}^A_\nu} \in \, g_{i,j}^{\pmb{P}_A}} \beta_{i,j}^{\mathcal{M}^A_{\nu}}. 
\end{equation} Meanwhile, the relevance parameters of interactions in the offspring models are inherited from their respective parent models.

\subsection{Instantiation of the Search Process}

Utilizing mutation, intra- and inter-population crossover mechanisms, we implement MCES as detailed in Algorithm~\ref{alg1}. MCES begins by randomly initializing eight distinct model populations (line 1). It then follows a series of iterative steps (lines 6-28), continuing until convergence. Each iteration involves optimizing offspring models and their relevance parameters within each population.

For every $\tau$ iterations, the algorithm (lines 9-16) selects and replaces the worst model $\mathcal{M}$ in each population $\pmb{P}$ based on the designated loss function (Eq.~(\ref{eq3})). The selection process can be expressed as: 
\begin{equation} \label{eq8} 
	\mathcal{M} = \arg \max _{\mathcal{M}_\nu \in \pmb{P}} \mathcal{L}( \mathcal{M}_\nu).
\end{equation} 
When the algorithm replaces the worst model $\mathcal{M}$ with the offspring model $\mathcal{M}'$, a new offspring $\mathcal{M}'$ is generated through intra-population crossover and subsequent mutation, enhancing genotypic diversity, thus enabling the search process to effectively avoid local optima and explore global regions.

For every $ep \times \tau$ iterations (lines 18-27),  we randomly select a pair of populations, $\pmb{P}_A$ and $\pmb{P}_B$, either based on a shared task (i.e., epilimnion or hypolimnion) across different lake types or on the same lake type but with different tasks, with selection probability balanced by a coin flip. 
This leads to the generation of new offspring models $\mathcal{M}_A'$, $\mathcal{M}_B'$ through inter-population crossover between $\pmb{P}_A$, $\pmb{P}_B$, promoting the exchange of advantageous genotypic patterns across different lake types and tasks. Meanwhile, each remaining population generates its offspring, $\mathcal{M}'$, through intra-population crossover, followed by mutation of all offspring.  

Finally, the algorithm concludes by delivering a set of the best models, one from each population (line 29),  thereby ensuring a comprehensive exploration and exploitation of the search space across diverse environmental contexts.

\begin{algorithm} [!t]
	\caption{Multi-population Cognitive Evolutionary Search}	
	\label{alg1}
	\textbf{Input}: Training dataset of four types of lakes $\pmb{L}_{\rm S}$, $\pmb{L}_{\rm M}$, $\pmb{L}_{\rm L}$, $\pmb{L}_{\rm xL}$, each lake has features $\pmb{f}_t$, simulated DO labels $\tilde{y}_{t}^{\rm epi}$, $\tilde{y}_{t}^{\rm hyp}$ over $T$ days; operation set  $\pmb{g}$. 
	\begin{algorithmic}[1] 
		\State Initialize eight populations $\pmb{P}^{\rm epi}_{\rm S}$, $\pmb{P}^{\rm hyp}_{\rm S}$, $\pmb{P}^{\rm epi}_{\rm M}$, $\pmb{P}^{\rm hyp}_{\rm M}$, $\pmb{P}^{\rm epi}_{\rm L}$, $\pmb{P}^{\rm hyp}_{\rm L}$, $\pmb{P}^{\rm epi}_{\rm xL}$, $\pmb{P}^{\rm hyp}_{\rm xL}$, of which any $\mathcal{M}$ has initialized $\pmb{\alpha}$ and $\pmb{\beta}$.
		\For{each $\pmb{P}$}
		\State Generate $\mathcal{M}'$ via intra-population crossover in $\pmb{P}$. 
		\Comment{Eq. (\ref{eq6})}
		\State Mutate $\mathcal{M}'$.
		\Comment{Eq. (\ref{eq5})}
		\EndFor
		\While{not converged} 
		\For{each $\pmb{P}$}
		\State Optimize $\mathcal{M}'$ with $\pmb{\alpha}'$ $\pmb{\beta}'$.
		\If{mod$(t, \tau)$ = 0}
		\State Select the worst $\mathcal{M}$.
		\Comment{Eq. (\ref{eq8})} 
		\State Replace $\mathcal{M}$ in $\pmb{P}$ with $\mathcal{M}'$.
		\If{mod$(t, ep \times \tau) \neq$ 0} 
		\State Generate $\mathcal{M}'$ via intra-population crossover. 
		\State Mutate $\mathcal{M}'$.
		\Comment{Eq. (\ref{eq5})} 
		\EndIf
		\EndIf
		\EndFor
		\If{mod$(t, ep \times \tau)$ = 0}
		\State Choose $(\pmb{P}_A, \pmb{P}_B)$ either by task or lake type.
		\State Generate $(\mathcal{M}_A', \mathcal{M}_B')$ via inter-population crossover of $\pmb{P}_A$, $\pmb{P}_B$.
		\Comment{Eq. (\ref{eq7})}
		\For{each $\pmb{P}$ not in $(\pmb{P}_A, \pmb{P}_B)$}
		\State Generate $\mathcal{M}'$ via intra-population crossover. 
		\EndFor
		\For{each $\pmb{P}$}
		\State Mutate $\mathcal{M}'$.
		\Comment{Eq. (\ref{eq5})}
		\EndFor
		\EndIf
		\EndWhile
		\State \Return the set of best models $\mathcal{M} = \arg \min _{\mathcal{M}_\nu \in \pmb{P}} \mathcal{L}( \mathcal{M}_\nu)$ from each population $\pmb{P}$. 
	\end{algorithmic}
\end{algorithm}

\section{Model Refinement}

Inspired by nature's replication and transcription processes, which translate genetic information into protein sequences to equip organisms with diverse functions, 
we proceed to a model refinement stage.  Here our objective is to refine the model to better leverage the features and interactions obtained from MCES. 
At this stage, we select the corresponding model by the lake type and task, and then use observed labels for the model refinement. 
Relevant features and interactions are selected according to their relevance parameters $\pmb{\alpha}$, $\pmb{\beta}$. If $\alpha_{i}=0$ or $\beta_{i,j}=0$, the corresponding features or interactions are fixed to be discarded permanently. 
Given the scarcity of observed data, we inherit parameters from the preceding LSTM  to ensure the model's effective learning, given as:  
\begin{equation} \label{eq9} 
	\begin{split}
		& \pmb{h}^{o}_t = {\rm LSTM} \Big([ \pmb{\alpha} \cdot \pmb{f}_t,  \pmb{\beta} \cdot \tilde{g}(\pmb{f}_t)];\pmb{h}^{o}_{t-1} \Big) \\
		& \hat{y}_t = \pmb{W}^{o} \cdot \pmb{h}^{o}_t + \pmb{b}^{o}
	\end{split} 
\end{equation} where $\pmb{h}^{o}_t$ represents a series of hidden states, with $\pmb{W}^{o}$ and $\pmb{b}^{o}$ as the weight and bias parameters. The relevance $\pmb{\alpha}$, $\pmb{\beta}$ are fixed and serve as attention units.

To address the disparity between abundant simulated and scarce observed labels, we've developed a new loss function for LSTM that integrates both types of data through weighted imputation~\cite{jia2019physics,wu2019deep,ye2022mane}. This assigns a greater weight to the loss on observed data and a smaller weight to simulated data, effectively addressing the scarcity of observed labels. The loss function is expressed as:
\begin{equation} \label{eq10}  
	\mathcal{L}(\mathcal{M}) = \frac{1}{|B|}\sum_{t \in B} \mathbb{I}\big(y_{t} \big) \Big(y_{t} -\hat{y}_{t} \Big) ^2
	+ \rho \Big(1-\mathbb{I}\big( y_{t} \big) \Big) \Big( \tilde{y}_{t} -\hat{y}_{t}   \Big) ^2,
\end{equation} where $\hat{y}_{t}$ denotes the predicted DO concentration, $y_{t}$ is the observed DO concentration, $ \tilde{y}_{t} $ is the simulated DO concentration, $\mathbb{I}(x)$ is an indicator function that equals $1$ if $x$ is observed (true) and $0$ otherwise (false), and $\rho$ is the tradeoff parameter balancing observed and simulated labels.

\section{Experimental Evaluation}

\subsection{Dataset}

We evaluate the proposed MCES for predicting DO concentration using a dataset that documents over 41 years of ecological observations from 375 lakes in the Midwestern USA, starting in 1979. This dataset includes around 5.58 million daily records, each with 39 fields of phenological features such as morphometric, flux data, weather conditions, trophic states, and land use details. 
Observed DO data were sourced from the Water Quality Portal (WQP). Lake residence time was taken from the HydroLAKES. Trophic state probabilities (eutrophic, oligotrophic, dystrophic) were from the dataset~\cite{meyer2024national}. Land use proportions of each lake's watershed were taken from the National Land Cover Database (NLCD). An account of these features is available in \href{https://zenodo.org/doi/10.5281/zenodo.10993058}{\textbf{Appendix C}}. For training MCES, we use data from all 375 lakes. We then selectively conduct testing on lakes that have the most comprehensive DO observations for each type. For large and extra-large lakes, we use data up to 2017 for training, 2018 for validation, and 2019 for testing. For small and medium lakes, where DO observations in 2019 are relatively sparse, we use data up to 2016 for training, 2017 for validation, and 2018 for testing.

\subsection{Baselines}

We compare to a set of baselines in our experiment: 
\textsl{Sim DO Conc.}: This baseline is the metabolic process-based model used in our first stage, leveraging few observed labels to calibrate simulations that can significantly augment the data for other baselines.
\textsl{LSTM}: As adopted in our model refinement stage,  LSTM incorporates simulated labels for weighted imputation and backward gradient adjustments, a necessity for convergence given the scarcity of observed labels.
\textsl{EA-LSTM \& KGSSL}~\cite{kratzert2019towards,ghosh2022robust}: These time series prediction models, which assimilate hydrological behavior and physical processes, respectively, are regarded as cutting-edge within hydrological and ecological domains. LSTM, EA-LSTM, and KGSSL use individual features for input without feature interaction modeling.
\textsl{AutoInt, AutoGroup, \& AutoFeature}~\cite{song2019autoint,liu2020autogroup,khawar2020autofeature}: These methods are at the forefront of feature interaction modeling. They have demonstrated their utility and versatility through extensive commercial deployment, showcasing their capacity to model complex feature combinations.

\newcolumntype{L}[1]{>{\raggedright\arraybackslash}p{#1}}
\newcolumntype{C}[1]{>{\centering\arraybackslash}p{#1}}
\newcolumntype{R}[1]{>{\raggedleft\arraybackslash}p{#1}}

\begin{table}[t]\scriptsize
	\centering
	\caption{Comparative performance of DO concentration ($g / m^{3}$) prediction in terms of root mean square error (RMSE) across different lake types and tasks. The mean and standard deviation (displayed in grey) of RMSE are calculated from five runs.}
	\begin{tabular}{L{1.78cm}C{1.18cm}C{1.18cm}C{1.18cm}C{1.18cm}C{1.18cm}C{1.18cm}C{1.18cm}C{1.18cm}}
		\toprule
		\multicolumn{1}{l}{\multirow{2}[3]{*}{Algo. Name}} & \multicolumn{2}{c}{Small lakes} & \multicolumn{2}{c}{Medium lakes} & \multicolumn{2}{c}{Large lakes} & \multicolumn{2}{c}{Extra-large lakes} \\
		\cmidrule(lr){2-3} \cmidrule(lr){4-5} \cmidrule(lr){6-7} \cmidrule(lr){8-9}
		& Epi. & Hyp. & Epi. & Hyp. & Epi. & Hyp. & Epi. & Hyp. \\
		\midrule
		\multirow{2}{*}{Sim DO conc.} & 1.943 & 2.212 & 1.940 & 2.217 & 2.620 & 2.937 & 1.536 & 2.772 \\
		& \textcolor{gray}{$\pm$ 0.000} & \textcolor{gray}{$\pm$ 0.000} & \textcolor{gray}{$\pm$ 0.000} & \textcolor{gray}{$\pm$ 0.000} & \textcolor{gray}{$\pm$ 0.000} & \textcolor{gray}{$\pm$ 0.000} & \textcolor{gray}{$\pm$ 0.000} & \textcolor{gray}{$\pm$ 0.000} \\
		\multirow{2}{*}{LSTM} & 1.802  & 1.973 & 1.744 & 2.001 & 2.298 & 2.630 & 1.479 & 2.594 \\
		& \textcolor{gray}{$\pm$ 0.079} & \textcolor{gray}{$\pm$ 0.064} & \textcolor{gray}{$\pm$ 0.092} & \textcolor{gray}{$\pm$ 0.081} & \textcolor{gray}{$\pm$ 0.088} & \textcolor{gray}{$\pm$ 0.043} & \textcolor{gray}{$\pm$ 0.068} & \textcolor{gray}{$\pm$ 0.056} \\
		\multirow{2}{*}{EA-LSTM} & 1.716 & 1.783 & 1.676 & 1.546 & 2.111 & 2.629 & 1.478 & 2.278  \\
		& \textcolor{gray}{$\pm$ 0.047} & \textcolor{gray}{$\pm$ 0.098} & \textcolor{gray}{$\pm$ 0.084} & \textcolor{gray}{$\pm$ 0.054} & \textcolor{gray}{$\pm$ 0.045} &  \textcolor{gray}{$\pm$ 0.043} & \textcolor{gray}{$\pm$ 0.039} & \textcolor{gray}{$\pm$ 0.062} \\
		\multirow{2}{*}{KGSSL} & 1.793 & 1.467 & 1.557 & 1.632 & 2.064 & 2.730 & 1.294 & 2.425 \\ 
		& \textcolor{gray}{$\pm$ 0.044} & \textcolor{gray}{$\pm$ 0.057} & \textcolor{gray}{$\pm$ 0.062} & \textcolor{gray}{$\pm$ 0.094} &  \textcolor{gray}{$\pm$ 0.103} &  \textcolor{gray}{$\pm$ 0.060} & \textcolor{gray}{$\pm$ 0.047} & \textcolor{gray}{$\pm$ 0.075} \\
		\multirow{2}{*}{AutoInt} & 1.510 & 1.406 & 1.516 & 1.626 & 1.716 & 1.924 & 1.112 & 1.847 \\
		&  \textcolor{gray}{$\pm$ 0.080} &  \textcolor{gray}{$\pm$ 0.097} & \textcolor{gray}{$\pm$ 0.088} & \textcolor{gray}{$\pm$ 0.094} & \textcolor{gray}{$\pm$ 0.072} & \textcolor{gray}{$\pm$ 0.078} & \textcolor{gray}{$\pm$ 0.081} & \textcolor{gray}{$\pm$ 0.085} \\
		\multirow{2}{*}{AutoGroup} & 1.473  & 1.509 & 1.364 & 1.875 & 1.384 & 1.600 & 0.937 & 1.953 \\
		& \textcolor{gray}{$\pm$ 0.078} & \textcolor{gray}{$\pm$ 0.080} & \textcolor{gray}{$\pm$ 0.059} & \textcolor{gray}{$\pm$ 0.072} & \textcolor{gray}{$\pm$ 0.075} & \textcolor{gray}{$\pm$ 0.068} &  \textcolor{gray}{$\pm$ 0.085} & \textcolor{gray}{$\pm$ 0.076} \\
		\multirow{2}{*}{AutoFeature} & 1.382 & 1.768 & 1.422 & 1.467 & 1.405 & 1.465 & 1.178 & 1.976  \\
		& \textcolor{gray}{$\pm$ 0.063}  &  \textcolor{gray}{$\pm$ 0.089} & \textcolor{gray}{$\pm$ 0.070} & \textcolor{gray}{$\pm$ 0.081} &  \textcolor{gray}{$\pm$ 0.084} & \textcolor{gray}{$\pm$ 0.082} & \textcolor{gray}{$\pm$ 0.078} & \textcolor{gray}{$\pm$ 0.093} \\
		\midrule
		\multirow{2}{*}{MCES (-refine)} & 1.851 & 2.044 & 1.812 & 2.024 & 2.374 & 2.775 & 1.530 & 2.648 \\
		& \textcolor{gray}{$\pm$ 0.204} & \textcolor{gray}{$\pm$ 0.210} & \textcolor{gray}{$\pm$ 0.227} & \textcolor{gray}{$\pm$ 0.223} & \textcolor{gray}{$\pm$ 0.215} & \textcolor{gray}{$\pm$ 0.218} & \textcolor{gray}{$\pm$ 0.208} & \textcolor{gray}{$\pm$ 0.243} \\
		\multirow{2}{*}{MCES (-multi)} & 1.315 & 1.624 & 1.390 & 1.473 & 1.390 & 1.572 & 1.151 & 1.848 \\
		& \textcolor{gray}{$\pm$ 0.182} & \textcolor{gray}{$\pm$ 0.207} & \textcolor{gray}{$\pm$ 0.203} & \textcolor{gray}{$\pm$ 0.204} & \textcolor{gray}{$\pm$ 0.208} & \textcolor{gray}{$\pm$ 0.194} & \textcolor{gray}{$\pm$ 0.197} & \textcolor{gray}{$\pm$ 0.235} \\
		\multirow{2}{*}{MCES (-inter)} & 1.107 & 1.375 & 1.161 & 1.354 & 1.004 & 1.307 & 1.040 & 1.536 \\
		& \textcolor{gray}{$\pm$ 0.179} & \textcolor{gray}{$\pm$ 0.193} & \textcolor{gray}{$\pm$ 0.192} & \textcolor{gray}{$\pm$ 0.190} & \textcolor{gray}{$\pm$ 0.198} & \textcolor{gray}{$\pm$ 0.189} & \textcolor{gray}{$\pm$ 0.194} & \textcolor{gray}{$\pm$ 0.223} \\
		\multirow{2}{*}{MCES} & 1.076 & 1.316 & 1.060 & 1.288 & 0.988 & 1.243 & 0.918 & 1.415 \\
		& \textcolor{gray}{$\pm$ 0.146} & \textcolor{gray}{$\pm$ 0.161} & \textcolor{gray}{$\pm$ 0.137} &  \textcolor{gray}{$\pm$ 0.159} & \textcolor{gray}{$\pm$ 0.169} & \textcolor{gray}{$\pm$ 0.156} & \textcolor{gray}{$\pm$ 0.171} & \textcolor{gray}{$\pm$ 0.215} \\ 
		\bottomrule
	\end{tabular}%
	\label{table01}
\end{table}

\subsection{Implementation Details.}

To implement MCES, we perform a grid search setting feature embedding size $|\pmb{f}^i_t|=15$. We use RDA optimizer~\cite{xiao2009dual,chao2019generalization} to discriminate the relevant and irrelevant feature interactions, with the learning rate $\gamma=10^{-3}$, adjustable hyperparameters $c=0.5, \mu=0.8$. We set the population size as $n=4$. We set the mutation mechanism as the mutation threshold $\lambda=0.2$, the mutation probability $\sigma=0.5$, and the mutation step size $\tau=10$. We set the inter-population crossover step size $ep=10$, and the tradeoff parameter $\rho=0.1$.
To assess the effectiveness of utilizing multiple populations and refining models, we conduct ablation studies with the following variants: \textsl{MCES (-refine)}: This variant drops the model refinement stage, using only the simulated labels from the first stage for training. \textsl{MCES (-multi)}: Instead of using multiple populations, this variant trains a single population on all data, and tests it separately across different lake types. This setup also excludes the inter-population crossover mechanism. \textsl{MCES (-inter)}: This variant does not include the inter-population crossover mechanism.

\begin{figure} [t]
	\includegraphics[width=1\linewidth]{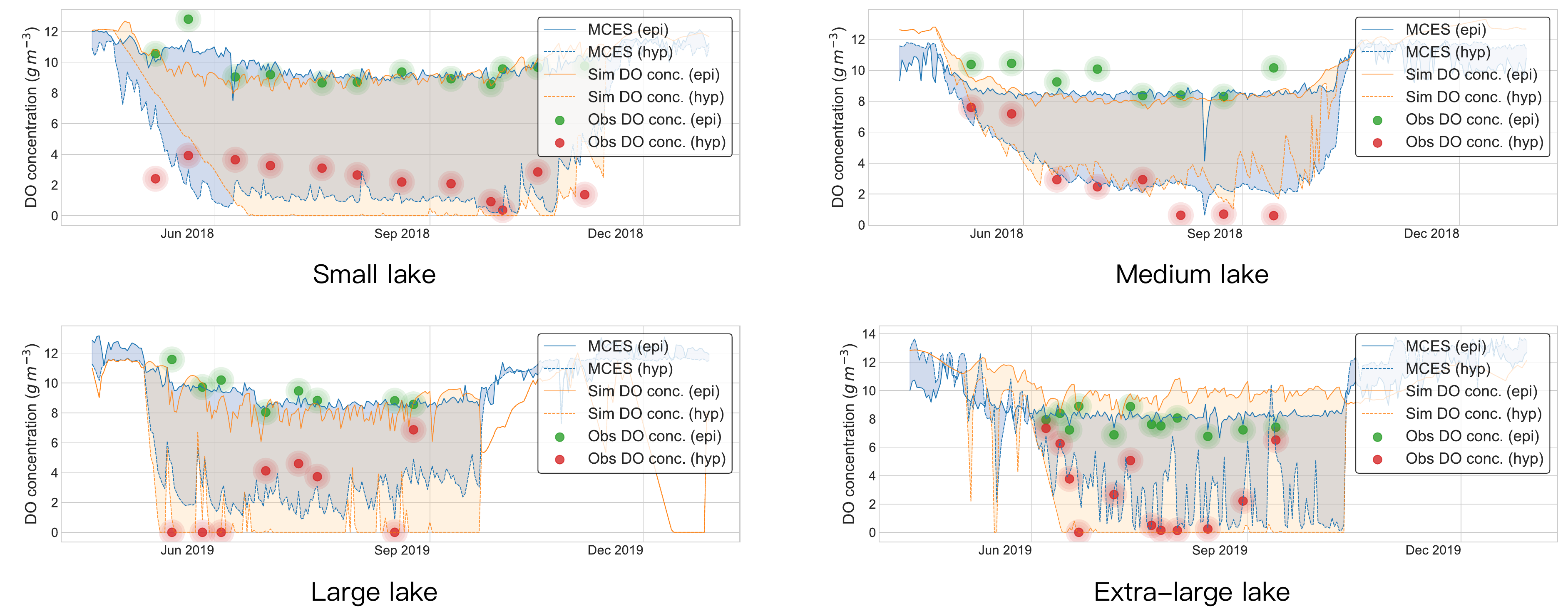}
	\caption{Time-series analysis of DO concentrations: a comparison of predicted (MCES), simulated, and observed values. } 
	\label{fig:3}
\end{figure} 

\subsection{Experimental Results} 

\subsubsection{Performance comparison.} Table~\ref{table01} presents a comparative analysis of MCES against various baselines, utilizing root mean square error (RMSE) across diverse lake types and tasks, with both mean and standard deviation calculated over five runs. From the results, we have the following key observations:
First, machine learning models universally outperform simulations alone, underscoring the value of integrating observed labels with simulated labels for enhanced prediction accuracy.
Second, EA-LSTM and KGSSL surpass LSTM in performance, evidencing the advantage of incorporating hydrological behaviors and physical processes into models, particularly when faced with a scarcity of labels.
Third, AutoInt, AutoGroup, and AutoFeature demonstrate the predictive power of feature interactions, offering significant improvements over models that rely solely on individual feature inputs.
Fourth, MCES outperforms all baseline models, attributing its success to the adaptive modeling of interactions through evolutionary operation selection. Unlike other methods that indiscriminately handle all feature interactions, MCES discerns relevant feature interactions for specific lake types and tasks, optimizing their impact while minimizing less pertinent ones.
Lastly, ablation study results reveal that MCES performs better than MCES (-refine), highlighting the value of combining real with simulated labels for accuracy. MCES also exceeds MCES (-multi), emphasizing the necessity of distinct populations for different ecological settings. Moreover, MCES outshines MCES (-inter), demonstrating the benefits of inter-population interactions. These interactions promote the sharing of effective feature interactions among populations, thereby improving the algorithm’s accuracy and stability. 

Figure~\ref{fig:3} offers a time-series comparison of predicted (i.e., MCES), simulated, and observed DO concentrations, with a specific emphasis on the summer season of the testing period. The analysis reveals that MCES predictions not only align closely with observed values but also demonstrate sensitivity to subtle features and interactions, enhancing their accuracy. While simulated DO concentrations also generally exhibit a clear trend, there are instances of slight deviation from observed data. Encouragingly, both predicted and simulated values largely mirror observed trends, highlighting the efficacy and significance of our proposed MCES in advancing research in this domain.

\begin{figure} [t]
	\centering{\includegraphics[width=0.9\textwidth]{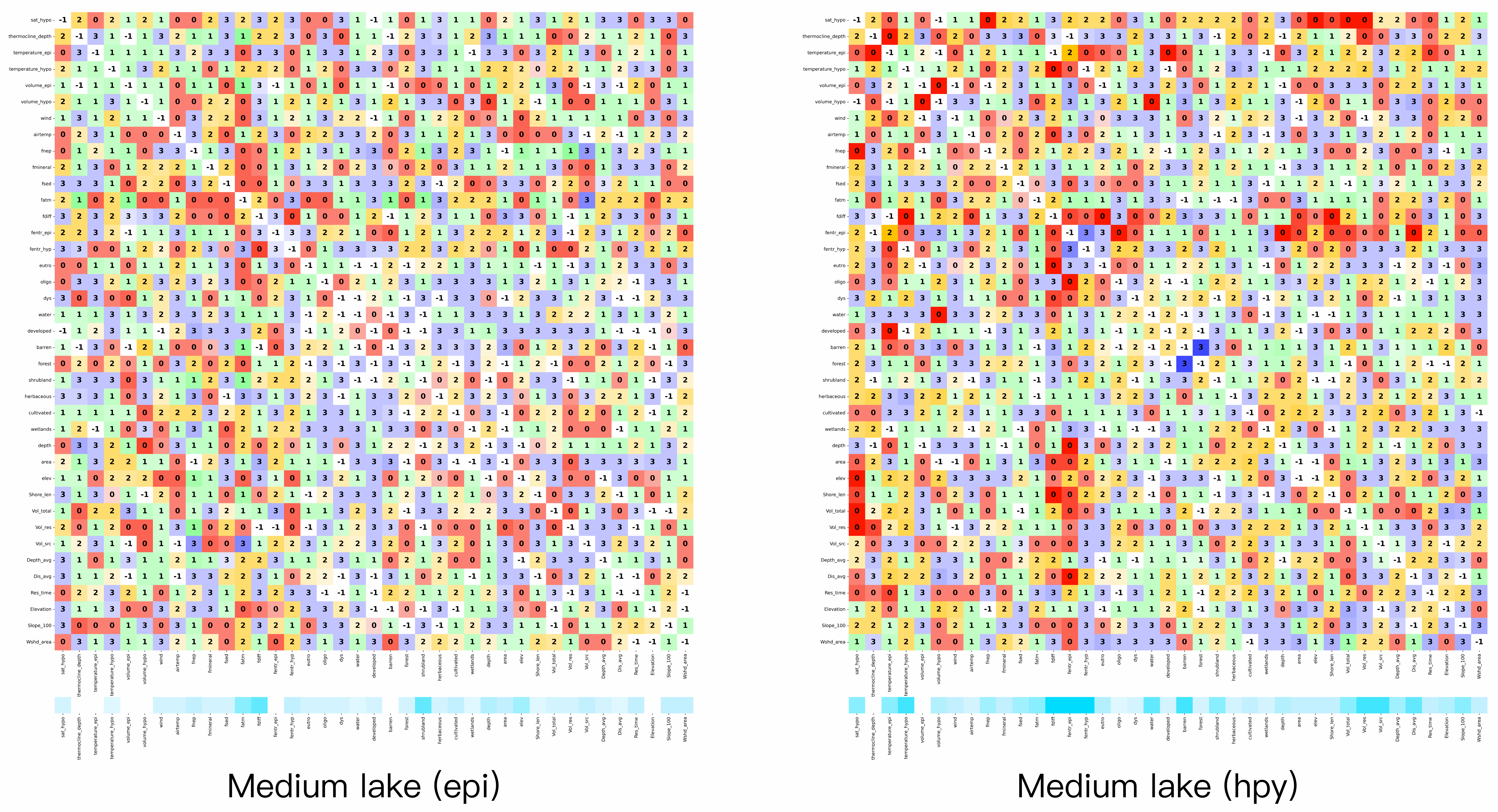}}
	\caption{ Visualization of gene maps for medium lakes. } 
	\label{fig:4}
\end{figure}

\subsubsection{Visualization of gene maps.} The model populations accommodate different lake types and tasks, leading to a rich diversity in model traits. These traits influence their survival and fitness rates, mirroring the selection process for operations and feature interactions. To demonstrate the model's evolutionary process and how feature interactions adapt across different lake types and tasks, we visualize the model's \textbf{gene maps}. Adopting an encoding where $\oplus=0, \otimes=1,  \boxplus=2, \boxtimes=3$, we can diagnose the model's fitness as a symmetric matrix. Distinct colors are allocated to each operation, creating a vibrant gene map where each gene symbolizes an interaction; like red ``$0$'', green ``$1$'', yellow ``$2$'', and blue ``$3$''. For example, a green ``$1$'' within the ``\verb|depth| $\times$ \verb|area|'' block signifies that the element-wise product~$\otimes$ is identified as the optimal operation for ``\verb|depth|'' to interact with ``\verb|area|''. The intensity of the colors on the gene map is directly correlated with the relevance of the interactions, with darker hues denoting higher relevance and lighter ones suggesting lesser importance. Individual features are also visually encoded as single-hued bars. Interactions deemed irrelevant, with their relevance parameters reduced to $0$, are excluded, leaving their corresponding genes depicted in white ``$-1$''.

In \href{https://zenodo.org/doi/10.5281/zenodo.10993058}{\textbf{Appendix D}}, we present gene maps for all lake types and tasks, using end-of-training data to highlight relevant feature interactions for DO concentration prediction. For instance, in Figure~\ref{fig:4}, we showcase gene maps for medium lakes. 

These maps reveal that, in larger lakes, the DO dynamics are predominantly influenced by sediment oxygen demand and atmospheric exchange, reflecting their extensive water volumes. Conversely, smaller lakes exhibit DO concentrations that are notably impacted by local land use and meteorological factors due to their shallower depths and greater vulnerability to changes in their external watershed environments. Across the board, temperature-related interactions are significant, affecting DO solubility and the lake's biological processes. Additionally, wind speed and atmospheric exchange flux stand out as key drivers of surface gas exchange influencing epilimnion, while the trophic state markers provide indicators of possible oxygen production in the epilimnion and eventual hypolimnetic depletion due to the formation of algal blooms. These findings suggest that diverse ecological factors interplay differently across lake environments, necessitating adaptable prediction models that can cater to these variances.

Additionally, we have included animations in the \href{https://zenodo.org/doi/10.5281/zenodo.10995166}{\textbf{supplementary}} materials to illustrate the evolution path of MCES. These animations display yearly changes in relevant feature interactions for DO concentrations from 1979. The enduring patterns of feature interactions hint at consistent ecological processes, while deviations in their relevance suggest adaptation to environmental shifts and human activities. Changes in the importance of certain interactions may stem from better land management or climate variations affecting lake stratification. Meanwhile, the emergence of new significant interactions could be a reaction to changes in lake usage or watershed practices. These temporal dynamics underscore the adaptability of MCES, which recalibrates the significance of feature interactions to align with the changing lake environments over time.


In \href{https://zenodo.org/doi/10.5281/zenodo.10993058}{\textbf{Appendix E}}, we also demonstrate the impact of feature interactions identified by MCES. By adjusting the RDA optimizer's parameters, we consistently choose a sparser set of feature interactions, accepting a trade-off in accuracy. Simultaneously, a random strategy is applied for comparative purposes, where operations for feature interactions are allocated at random, and some interactions are arbitrarily removed as sparsity intensifies. The gene map showcased at a feature interaction sparsity level around $0.5$ offers insight into the model's structure under reduced complexity. This experiment highlights MCES's superior performance even as many feature interactions are discarded, emphasizing its precision in identifying relevant interactions under task guidance. Conversely, the random approach shows a quicker performance drop due to the loss of important interactions. When feature interactions become exceedingly sparse, both methodologies suffer in performance, indicating that a limited set of feature interactions fails to significantly contribute to the model's predictive capabilities. In such cases, individual features primarily drive performance.

\section{Conclusion}

This paper presents a novel evolutionary algorithm, namely \textsl{Multi-population Cognitive Evolutionary Search (MCES)}, blending adaptive learning with natural processes for predicting DO concentrations. MCES employs multi-population models customized for varied lake types and tasks, reflecting the diverse survival strategies of species across various habitats. Evaluated on a variety of lakes in the Midwestern USA, MCES not only demonstrates accurate DO concentration predictions with limited observed data but also reveals sophisticated phenological patterns, highlighting its utility for environmental science and freshwater management. MCES introduces an innovative concept of ``AI from nature, for nature''. We encourage further exploration and development of new algorithms inspired by MCES, aimed at benefiting the environment.

\subsubsection{\ackname} This research was partially supported by the University of Pittsburgh Center for Research Computing through the resources provided.

%
%
%
\bibliographystyle{splncs04}
\bibliography{mybibliography}
%
%
%
%
%
\end{document}